\documentclass[10pt,twocolumn,letterpaper]{article}

\usepackage{cvpr}      

\usepackage{graphicx}
\usepackage{amsmath}
\usepackage{amssymb}
\usepackage{booktabs}

\usepackage{bbding}
\usepackage{pifont}
\usepackage{wasysym}

\usepackage[pagebackref,breaklinks,colorlinks]{hyperref}
\usepackage{xcolor}

\usepackage[capitalize]{cleveref}
\crefname{section}{Sec.}{Secs.}
\Crefname{section}{Section}{Sections}
\Crefname{table}{Table}{Tables}
\crefname{table}{Tab.}{Tabs.}

\def\cvprPaperID{3731} 
\def\confName{CVPR}
\def\confYear{2023}

\begin{document}

\title{CLAMP: Prompt-based Contrastive Learning for Connecting Language and Animal Pose}

\author{
{Xu Zhang\textsuperscript{1} \quad Wen Wang\textsuperscript{2} \quad Zhe Chen\textsuperscript{1}  \quad Yufei Xu\textsuperscript{1} \quad Jing Zhang\textsuperscript{1} \quad Dacheng Tao\textsuperscript{1}} \\
{\normalsize \textsuperscript{1}The University of Sydney, Australia}
\quad
{\normalsize \textsuperscript{2}Zhejiang University, China} \\
{\tt\footnotesize \{xzha0930,yuxu7116\}@uni.sydney.edu.au \ wwen@zju.edu.cn} \\
{\tt\footnotesize \{zhe.chen1,jing.zhang1\}@sydney.edu.au \ dacheng.tao@gmail.com}
}

\maketitle

\begin{abstract}
Animal pose estimation is challenging for existing image-based methods because of limited training data and large intra- and inter-species variances. Motivated by the progress of visual-language research, we propose that pre-trained language models (\eg, CLIP) can facilitate animal pose estimation by providing rich prior knowledge for describing animal keypoints in text. However, we found that building effective connections between pre-trained language models and visual animal keypoints is non-trivial since the gap between text-based descriptions and keypoint-based visual features about animal pose can be significant. To address this issue, we introduce a novel prompt-based \textbf{C}ontrastive learning scheme for connecting \textbf{L}anguage and \textbf{A}ni\textbf{M}al \textbf{P}ose (CLAMP) effectively. The CLAMP attempts to bridge the gap by adapting the text prompts to the animal keypoints during network training. The adaptation is decomposed into spatial-aware and feature-aware processes, and two novel contrastive losses are devised correspondingly. In practice, the CLAMP enables the first cross-modal animal pose estimation paradigm. Experimental results show that our method achieves state-of-the-art performance under the supervised, few-shot, and zero-shot settings, outperforming image-based methods by a large margin. The code is available at https://github.com/xuzhang1199/CLAMP.

\end{abstract}

\section{Introduction}
\label{sec:intro}

Animal pose estimation aims to locate and identify a series of animal body keypoints from an input image. It plays a key role in animal behavior understanding, zoology, and wildlife conservation which can help study and protect animals better. 
Although the animal pose estimation task is analogous to human pose estimation~\cite{andriluka20142d} to some extent, we argue that the two tasks are very different. For example, animal pose estimation involves multiple animal species, while human pose estimation only focuses on one category. Besides, it is much more difficult to collect and annotate animal pose data covering different animal species, thus existing animal pose datasets are several times smaller than the human pose datasets \cite{coco} regarding the number of samples per species. Recently, Yu \etal \cite{ap-10k} attempted to alleviate this problem by presenting the largest animal pose estimation dataset, \ie, AP-10K, which contains 10K images from 23 animal families and 54 species and provides the baseline performance of SimpleBaseline~\cite{xiao2018simple} and HRNet~\cite{wang2020deep}. Despite this progress, the volume of this dataset is still far smaller than the popular human pose dataset, such as MS COCO\cite{coco} with 200K images. 

\begin{figure}[t]
  \centering
  \includegraphics[width=1.0\linewidth]{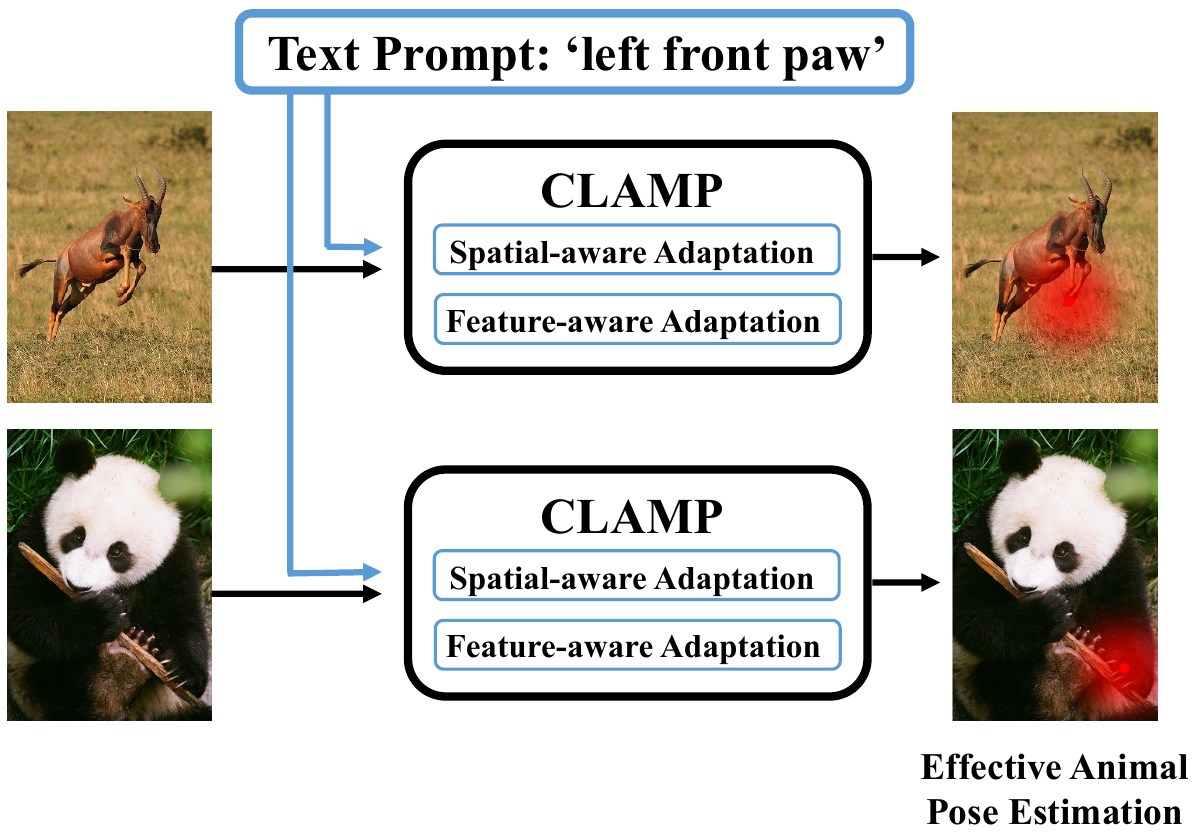}
  \caption{
  Conceptualized visualization of our CLAMP method. Regarding the animal pose estimation task, we proposed to exploit rich language information from texts to facilitate the visual identification of animal keypoints. To better connect texts and animal images, we devise the CLAMP to adapt pre-trained language models via a spatial-aware and a feature-aware process. As a result, the CLAMP helps deliver better animal pose estimation performance.
  }
   \label{fig:title}
\end{figure}

With diverse species and limited data, current animal pose datasets usually have large variances in animal poses which include both intra-species and inter-species variances. More specifically, the same animal can have diverse poses, \eg, pandas can have poses like standing, crawling, sitting, and lying down. Besides, the difference in the poses of different animal species can also be significant, \eg, horses usually lie down to the ground, while monkeys can be in various poses. Furthermore, even with the same pose, different animals would have different appearances. As an example, the joints of monkeys are wrinkled and hairy, while those of hippos are smooth and hairless. As a result, it could be extremely challenging for current human pose estimation methods to perform well on the animal pose estimation task without sufficient training data. Although image-based pre-training methodologies can be helpful in mitigating the problem of insufficient data, the huge gap between the pre-training datasets (\eg ImageNet~\cite{imagenet} image classification dataset and MS COCO human pose dataset~\cite{coco}) and the animal pose datasets could compromise the benefits of pre-training procedures.

Rather than only using images to pre-train models, we notice that the keypoints of different poses and different animals share the same description in natural languages, thus the language-based pre-trained models can be beneficial to compensate for the shortage of animal image data. For example, if a pre-trained language model provides a text prompt of ``a photo of the nose", we can already use it to identify the presence of the nose keypoint in the image without involving too much training on the new dataset. Fortunately, a recently proposed Contrastive Language-Image Pre-training (CLIP) \cite{clip} model can provide a powerful mapping function to pair the image with texts effectively. Nevertheless, we found that fine-tuning the CLIP on the animal pose dataset could still suffer from large gaps between the language and the images depicting animals. In particular, the vanilla CLIP model only learns to provide a text prompt with general language to describe the entire image, while the animal pose estimation requires pose-specific descriptions to identify several different keypoints with their locations estimated from the same image. To this end, it is important to adapt the pre-trained language model to the animal pose dataset and effectively exploit the rich language knowledge for animal pose estimation.

To address the above issue, we propose a novel prompt-based contrastive learning scheme for effectively connecting language and animal pose (called CLAMP), enabling the first cross-modal animal pose estimation paradigm. In particular, we design pose-specific text prompts to describe different animal keypoints, which will be further embedded using the language model with rich prior knowledge. By adapting the pose-specific text prompts to visual animal keypoints, we can effectively utilize the knowledge from the pre-trained language model for the challenging animal pose estimation. However, there is a significant gap between the pre-trained CLIP model (which generally depicts the entire image) and the animal pose task (which requires the specific keypoint feature discriminative to and aligned with given text descriptions). To this end, we decompose the complicated adaptation into a spatial and feature-aware process. Specifically, we devise a spatial-level contrastive loss to help establish spatial connections between text prompts and the image features. A feature-level contrastive loss is also devised to make the visual features and embedded prompts of different keypoints more discriminative to each other and align their semantics in a compatible multi-modal embedding space. 
With the help of the decomposed adaptation, effective connections between the pre-trained language model and visual animal poses are established. Such connections with rich prior language knowledge can help deliver better animal pose prediction.

In summary, the contribution of this paper is threefold:
\begin{itemize}
\item We propose a novel cross-modal animal pose estimation paradigm named CLAMP to effectively exploit prior language knowledge from the pre-trained language model for better animal pose estimation.
\item We propose to decompose the cross-modal adaptation into a spatial-aware process and a feature-aware process with carefully designed losses, which could effectively align the language and visual features.
\item Experiments on two challenging datasets in three settings, \ie, 1) AP-10K~\cite{ap-10k} dataset (supervised learning, few-shot learning, and zero-shot learning) and 2) Animal-Pose~\cite{cao2019cross} dataset (supervised learning), validate the effectiveness of the CLAMP method.
\end{itemize}

\section{Related work}
\subsection{Pose estimation}
Pose estimation is a challenging and active research area in computer vision. Most of the existing methods~\cite{fang2017rmpe,newell2016stacked,liu2020improving,wang2020deep,xiao2018simple,zhang2021towards,cao2017realtime,newell2017associative,geng2021bottom,xu2022vitpose,zhang2021towards} focus on human pose estimation and simply predict the locations of keypoints based on images. Although they obtain superior performance for human pose estimation, these methods face difficulties in generalizing to animal pose estimation tasks, where there are large intra- and inter-species variances for different animal instances and limited training data per species. To tackle this problem, previous methods resort to domain adaptation or knowledge distillation~\cite{jing2021amalgamating,yang2022deep,yang2022factorizing} for animal pose estimation~\cite{mu2020learning,li2021synthetic,sanakoyeu2020transferring,cao2019cross,ap-10k}. For example, Cao \etal~\cite{cao2019cross} propose a cross-domain adaptation method, which transfers the knowledge in labeled human pose data to handle the unlabeled animal pose data. Li \etal~\cite{li2021synthetic} carry out the domain adaptation from synthetic to real data for animal pose estimation. Recently, Yu \etal~\cite{ap-10k} validate the effectiveness of leveraging pose estimation models pre-trained on human datasets for fine-tuning. However, they only focus on the knowledge of image modality and still struggle to deal with multiple animal species with large variances in appearance, texture, and pose, especially in settings of limited data. 

In this paper, we try to address this problem from a novel perceptive, \ie, using rich prior knowledge of language modality. We argue that although the features of keypoints from different animal images may have large variances, they share the same description in languages. Motivated by this, we propose a novel prompt-based contrastive learning scheme with decomposed spatial-aware and feature-aware adaptation processes for effectively connecting language and animal pose.

\subsection{Vision-language models}
Vision-language models cover a wide range of research topics~\cite{xu2015show,antol2015vqa,anderson2018vision,qiao2019mirrorgan,rim}, while we focus on reviewing the most related works on vision-language pre-training and fine-tuning. 
Vision-language pre-training has witnessed significant progress in the last few years, which generally learns an image encoder and a text encoder jointly~\cite{clip,align,cloob,declip}. A representative work is contrastive language–image pre-training dubbed CLIP~\cite{clip}, which uses 400 million text-image paired data to pre-train a multi-modal model. Experiments show that CLIP can help achieve effective few-shot or even zero-shot classification by simply exploring the relations between text features and image features.

Although significant progress has been made in vision-language pre-training, how to effectively adapt these pre-trained models to downstream tasks is still challenging and actively studied. For example, CoOp~\cite{coop} and CoCoOp~\cite{cocoop} take inspiration from prompt learning in NLP~\cite{liu2021pre} and propose to utilize learnable text embedding for better image classification. Similarly, CLIP-adaptor~\cite{clip-adaptor} and TIP-adaptor~\cite{tip-adaptor} improve the model performance on downstream tasks through a lightweight adaptor. While the above methods focus on adapting CLIP for the image classification task, DenseCLIP~\cite{denseclip} proposes a language-guided fine-tuning method for applying the pre-trained models to semantic segmentation and instance segmentation. GLIP~\cite{li2022grounded} studies how to use image-text pairs to obtain a well pre-trained model that is suitable for object detection and phrase grounding. Different from them, our CLAMP makes the first attempt to leverage the language knowledge from the vision-language pre-trained model for animal pose estimation via specially designed pose-specific prompts and decomposed adaptation in both spatial and feature levels.


\section{Method}
\subsection{Preliminary}
\textbf{Animal pose estimation pipeline} Similar to the human pose estimation task, animal pose estimation aims to locate $N$ keypoints of each animal instance in the input image. We follow most of the existing pose estimation methods and apply a typical top-down keypoint detection pipeline~\cite{xiao2018simple,wang2020deep}, 
\ie, firstly using a detector to detect all animal instances in the image, then detecting the keypoints for each instance. Specifically, the heatmap representation is usually used to denote the location of each keypoint. 
We denote the cropped instance image as $I\in\mathbb{R}^{h\times w\times 3}$, where $h$ and $w$ are the height and width of the image, respectively. The image encoder $f_{extr}$ extracts the image feature $F\in\mathbb{R}^{h_0\times w_0\times C}$ from $I$, where $h_0$, $w_0$, and $C$ are the height, width, and the number of channels of the extracted feature, respectively. An ImageNet~\cite{imagenet} pre-trained backbone network, \eg, ResNet-50~\cite{resnet} or HRNet-32~\cite{wang2020deep}, is usually employed as the image encoder in image-based methods, while we adopt CLIP pre-trained ResNet-50 and ViT~\cite{vit} in our CLAMP to leverage the language knowledge. A typical ratio $s_0$ between $h$ and $h_0$ is 32 for ResNet-50 and 4 for HRNet-32. Then, the keypoint predictor $f_{pred}$ decodes $F$ into a heatmap $H\in\mathbb{R}^{h_1\times w_1\times N}$, which typically consists of several deconvolution layers depending on the ratio $s_0$ and a convolutional prediction layer. The ratio $s_1$ between $h$ and $h_1$ is 4. Finally, we get the coordinates of $N$ keypoints $K\in \mathbb{R}^{N\times 2}$ by applying a simple argmax operation on each heatmap and multiplying the coordinates with the scale ratio $s_1$ to recover to the original scale:
\begin{equation} \small
    K_n = s_1 \cdot \underset{1\leq i \leq h_1,1 \leq j \leq w_1}{argmax}H_n(i,j), \quad n=1,...,N,
\end{equation}
where $K_n$ is the 2D coordinate of $n$-th keypoint and $H_n$ is $n$-th heatmap in $H$. 

\begin{figure*}[t]
  \centering
  \includegraphics[width=1.0\linewidth]{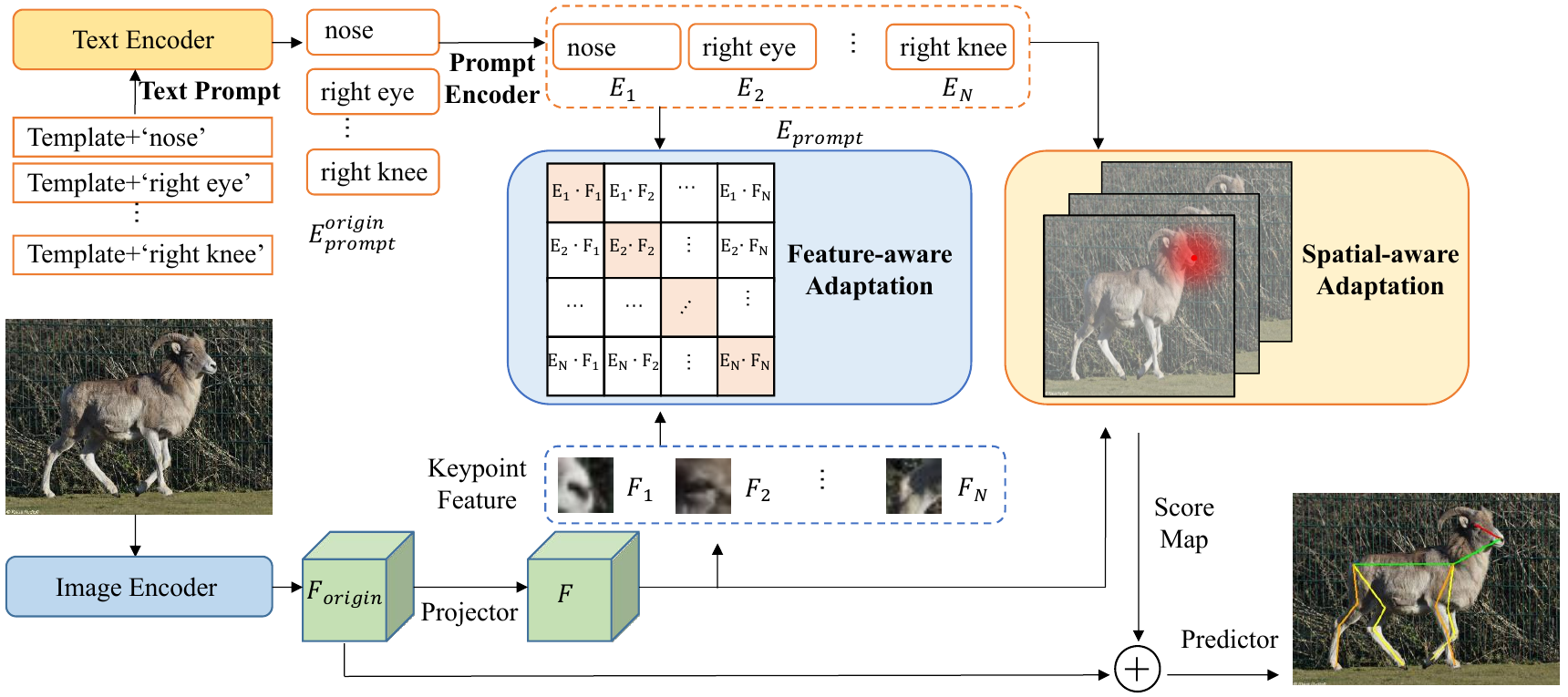}
  \caption{Conceptualized illustration of the proposed CLAMP method. The pipeline contains an image encoder extracting the visual feature of the input image, a text encoder encoding the text prompt with rich prior knowledge, and two adaptation modules for adapting prior language knowledge to visual animal pose. The obtained presence score map can help deliver effective animal pose estimation.
  }
   \label{fig:pipeline}
\end{figure*}

\textbf{Language model} CLIP~\cite{clip} provides visual-feature-compatible pre-trained language models by leveraging a large number of paired images and text descriptions in pre-training. By using natural language to reference learned visual concepts, CLIP enhances the generality of the pre-trained models, enabling effective knowledge transfer to downstream tasks. For example in classification, it uses ``A photo of a/an \{object\}'' as a template to formulate text prompts, where \{object\} can be filled with names of different categories. By calculating the similarity between image features and different embedded text prompts, the CLIP pre-trained models can directly adapt to the classification task because it is similar to the pre-training process.
CLIP aligns visual concepts with languages and demonstrates effective knowledge adaptation to downstream tasks. Motivated by this, we propose to exploit the prior language knowledge in CLIP pre-trained models and design a novel cross-modal animal pose estimation paradigm named CLAMP.

\subsection{CLAMP}
Our CLAMP includes the introduction of a set of pose-specific text prompts and two decomposed adaptation processes for leveraging prior language knowledge. Fig. \ref{fig:pipeline} illustrates the proposed CLAMP method.

\subsubsection{Pose-specific text prompts}
Different from the classification tasks explored in CLIP, pose estimation is agnostic to the category of the animal instance and needs to find the position of a set of local keypoints in each image. Thus, we need to design pose-specific text prompts for animal pose estimation. Motivated by CoOp~\cite{coop}, we use a learnable template of $k$ learnable prefix tokens instead of the fixed ``A photo of a'' template in CLIP to learn a prompt template that adapts better to pose estimation. Accordingly, we fill in \{object\} with the names of different keypoints like `nose' to get $N$ text prompts, \ie,
\begin{equation} \small
    p_n = [T]_1[T]_2\ldots[T]_k[KeyPoint]_n, \quad n=1,...,N,
    \label{eq:prompt}
\end{equation}
where $[T]_i, i\in \{1,2,\ldots, k\}$ represents the learnable prefix tokens, $[KeyPoint]_n$ represents the $n$-th keypoint name, and $N$ is the number of keypoints. The $N$ text prompts are mapped to the multi-modal embedding space by using a CLIP pre-trained text encoder to get the prompt embedding $E_{prompt}^{origin}\in\mathbb{R}^{N\times C_{emb}}$. Considering the intrinsic relationship between different keypoints is important for pose estimation, we further employ a lightweight prompt encoder (\ie, a single transformer layer) to model the relationship between the prompt embeddings of different keypoints and promote their interactions. After that, cross attention is applied to enhance the prompt embeddings with the image feature~\cite{denseclip}, generating the enhanced prompt embedding $E_{prompt}\in\mathbb{R}^{N\times C_{emb}}$.

\subsubsection{Pose-aware language knowledge adaptation}

Although CLIP exhibits effective knowledge transfer ability in downstream classification with the help of its designed prompts, it is still challenging to directly adapt the text prompts to animal pose estimation due to the lack of spatial connection between text description and image feature in CLIP pre-training. To address this challenge, we decompose the cross-modal adaptation into a spatial-aware process and a feature-aware process. Furthermore, a spatial-level contrastive loss and a feature-level contrastive loss are devised to constrain the two processes, respectively.

\textbf{Spatial-aware adaptation} Spatial-aware adaptation aims at establishing spatial connections between the text prompts and image features, which can provide positional information for the animal pose. In this process, we devise a spatial-level contrastive loss to query the possibility of the presence of different animal keypoints in spatial dimension. Specifically, we feed the input image into the image encoder to obtain the image feature $F_{origin}\in\mathbb{R}^{H\times W\times C}$, where $H$, $W$, and $C$ represent the height, width, and the number of channels, respectively. Then the obtained $F_{origin}$ is mapped to the multi-modal embedding space through a projector, obtaining $F\in\mathbb{R}^{H\times W\times C_{emb}}$. For different image encoders, we employ slightly different projectors. For example, if the encoder takes the form of a ViT~\cite{vit}, the projector is a linear projection layer with the cls token as input. For the ResNet~\cite{resnet} encoder, the projector contains a global average pooling layer, a multi-head self-attention (MHSA) layer, and a linear projection layer, following the design in CLIP~\cite{clip}. After that, the extracted prompt embeddings and image features are normalized and used to query the presence possibility of different keypoints via the inner product. Thus, we can get the presence score at each spatial position:
\begin{equation} \small
\begin{aligned}
   S_{ijn} &= F_{ij} \cdot E_{prompt}^n, \\ i&=1,...,H; j=1,...,W; n=1,...,N,
\end{aligned}
   \label{eq:pixel-1}
\end{equation}
where $F_{ij}\in\mathbb{R}^{1\times C_{emb}}$ is the feature vector of $F$ at pixel $(i, j)$, and $E_{prompt}^n$ is the $n$-th prompt embedding in $E_{prompt}$. Accordingly, we can get the presence score map by collecting and stacking the scores at all locations and prompts, \ie,
\begin{equation} \small
        S = Stack(S_{ijn}), 
        \quad S \in\mathbb{R}^{H\times W\times N},
        \label{eq:pixel-2}
\end{equation}
where $Stack$ represents the stack operation.

Considering the effectiveness of Gaussian heatmap in pose estimation~\cite{xiao2018simple,wang2020deep}, we use the target 2D Gaussian heatmap $H_{target}$ to supervise the estimated score map via the following spatial-level contrastive loss, \ie,
\begin{equation} \small
    \mathcal{L}_{spatial} = MSE(Upsample(S), H_{target}),
\end{equation}
where $MSE$ is the mean squared error loss. We upsample the score map to align the spatial size of the score map and the target heatmap.

\textbf{Feature-aware adaptation} Given CLIP pre-training only learns to reference the entire image while animal pose requires more discriminative keypoint features to align with the corresponding text prompts, we introduce a feature-level contrastive loss for feature-aware adaptation. We encourage the visual feature of a specific keypoint to be close to the text prompt describing the corresponding keypoint and to be far away from those describing other keypoints, and apply the same operation on embedded text prompts, thereby enhancing the discriminative ability of the extracted text and image features and facilitating their alignment. Specifically, during the training process, we use the ground truth locations of the keypoint $K\in\mathbb{R}^{N\times 2}$ to perform grid sampling on $F$ to obtain the local visual features of $N$ keypoints, \ie, $F_{n}\in\mathbb{R}^{1\times C_{emb}}, n=1,...,N$. The stacked keypoint feature can be obtained, \ie,
\begin{equation} \small
    F_{keypoint} = Stack(F_{n}),
    \quad F_{keypoint} \in\mathbb{R}^{N\times C_{emb}},
    \label{eq:gridsampling}
\end{equation}
where $Stack$ represents the stack operation. Then, we calculate the semantic matching score map between the visual feature of keypoints and prompt embeddings as follows:
\begin{equation} \small
    M = \hat{F}_{keypoint}  \hat{E}_{prompt}^T,
    \quad M \in\mathbb{R}^{N\times N},
    \label{eq:sementic-contrast}
\end{equation}
where $\hat{F}_{keypoint}$ and $\hat{E}_{prompt}$ are normalized keypoint features and prompt embeddings, respectively. 
Since there is only one prompt embedding describing one given visual keypoint feature, we can simply use the diagonal matrix as the matching target $M_{label}$.
We perform contrastive learning on both prompt embeddings and keypoint features based on the following feature-level contrastive loss, \ie,
\begin{equation} \small
    \mathcal{L}_{feature} = \frac{1}{2} (CE(M, M_{label}) + CE(M^T, M_{label})),
\end{equation}
where $CE$ represents the cross entropy loss.

\subsubsection{Final Prediction and Learning Objective}

With the designed pose-specific text prompts and the decomposed cross-modal adaptation, our proposed CLAMP could connect text descriptions to visual features, making it possible to adapt the rich prior language knowledge from pre-trained language models to animal pose estimation.
To let language knowledge collaborate with image features for animal pose estimation, we fuse the image features and the spatial presence score maps, \ie,
\begin{equation} \small
    F_{fuse} = F_{origin} \oplus S,
    \quad F_{fuse} \in\mathbb{R}^{H\times W\times (C+N)},
\end{equation}
where $\oplus$ represents the concatenate operation along the channel dimension. Then, $F_{fuse}$ is fed into a keypoint predictor to predict the pose heatmap. The prediction results are supervised by a prediction loss $\mathcal{L}_{pred}$, which adopts the form of $MSE$ loss between the predicted heatmap and ground truth heatmap. The overall training loss can be written as:
\begin{equation} \small
    \mathcal{L}_{total} = \mathcal{L}_{pred} + \alpha_1 \cdot \mathcal{L}_{spatial} + \alpha_2 \cdot \mathcal{L}_{feature},
\end{equation}
where $\alpha_1$ and $\alpha_2$ are two hyper-parameters to balance the importance of $\mathcal{L}_{spatial}$ and $\mathcal{L}_{feature}$.


\section{Experiments}
\subsection{Experimental setup}
\label{subsec:experimentsetting}

\textbf{Datasets and evaluation metrics} We employ the AP-10K~\cite{ap-10k} and Animal-Pose~\cite{cao2019cross} datasets to evaluate the performance of the proposed CLAMP method. The AP-10K dataset contains 10,015 images collected and filtered from 23 animal families and 54 species, which is the largest and most diverse dataset for animal pose estimation. 17 keypoints are annotated in the dataset, \ie, two eyes, one nose, one neck, two shoulders, two elbows, two knees, two hips, four paws, and one tail. We adopt the training set during the training process and evaluate the model's performance on the validation set. On the other hand, the Animal-Pose dataset covers 5 different animal species with over 4,000 images. 20 keypoints are annotated in each animal instance, including two eyes, throat, nose, withers, two earbases, one base of the tail, four elbows, four knees, and four paws. Similarly, we adopt the training set for training and report the results on the validation set. Following the common practice in animal pose estimation, we adopt the average precision (AP) as the main metric on the two datasets, which is computed based on the object keypoint similarity (OKS). The detailed protocol definitions can be found in \cite{xiao2018simple}.

\begin{table*}[t]
  \footnotesize
  \centering
  \begin{tabular}{c|c|c|ccc ccc}
    \hline
    Method & Backbone & Pre-train & $AP$ & $AP_{50}$ & $AP_{75}$ & $AP_{M}$ & $AP_{L}$ & $AR$ \\
    \hline
    SimpleBaseline~\cite{xiao2018simple} & ResNet-50 & ImageNet & 70.2 & 94.2 & 76.0 & 45.5 & 70.4 & 73.5 \\
    SimpleBaseline~\cite{xiao2018simple} & ResNet-50 & CLIP     & 70.9 & 94.6 & 76.8 & 44.8 & 71.2 & 74.1\\
    CLAMP (ours)     & ResNet-50 & CLIP     & 72.9 & 95.4 & 79.4 & 43.2 & 73.2 & 76.3 \\
    \hline
    SimpleBaseline~\cite{xiao2018simple}   & ViT-Base & CLIP     & 72.6 & 94.7 & 79.5 & 43.4 & 72.8 & 75.8\\
    CLAMP (ours)      & ViT-Base & CLIP     & 74.3 & 95.8 & 81.4 & 47.6 & 74.9 & 77.5\\
    \hline
  \end{tabular}
  \caption{Performance comparison on AP-10K~\cite{ap-10k}.}
  \label{overall-ap10k}
\end{table*}

\begin{table*}[t]
  \footnotesize
  \centering
  \begin{tabular}{c|c|c|ccc ccc}
    \hline
    Method & Backbone & Pre-train & $AP$ & $AP_{50}$ & $AP_{75}$ & $AP_{M}$ & $AP_{L}$ & $AR$ \\
    \hline
    SimpleBaseline~\cite{xiao2018simple} & ResNet-50 & ImageNet & 51.1 & 85.4 & 50.2 & 26.7 & 51.3 & 56.3 \\
    SimpleBaseline~\cite{xiao2018simple} & ResNet-50 & CLIP     & 51.8 & 84.3 & 52.5 & 26.1 & 52.0 & 56.9\\
    CLAMP (ours)  & ResNet-50 & CLIP     & 54.0 & 85.9 & 56.2 & 26.4 & 54.2 & 58.9\\
    \hline
    SimpleBaseline~\cite{xiao2018simple}   & ViT-Base & CLIP     & 57.4 & 88.9 & 61.2 & 20.9 & 57.8 & 61.2\\
    CLAMP   (ours)    & ViT-Base & CLIP     & 61.2 & 91.3 & 64.7 & 36.8 & 61.7 & 65.2\\
    \hline
  \end{tabular}
  \caption{20-shot performance comparison on AP-10K~\cite{ap-10k}.}
  \label{ap10k-20shot}
\end{table*}


\begin{table*}[t]
  \centering
  \footnotesize
  \begin{tabular}{c|c|c|c|ccc ccc}
    \hline
    Method & Backbone & Train & Test & $AP$ & $AP_{50}$ & $AP_{75}$ & $AP_{M}$ & $AP_{L}$ & $AR$ \\
    \hline
    SimpleBaseline~\cite{xiao2018simple} & ResNet-50 & Bovidae & Canidae & 41.3 &	79.4 & 36.4	& 26.8 & 41.3 &	49.1\\
    CLAMP (ours)  & ResNet-50 & Bovidae & Canidae & 46.9 &	84.4 & 45.6 & 30.3 & 46.9 &	53.8\\
    \hline
    SimpleBaseline~\cite{xiao2018simple}   & ResNet-50 & Canidae & Felidae & 39.6 &	74.1 & 34.5 & 9.5 & 40.2 & 46.6\\
    CLAMP   (ours)    & ResNet-50 & Canidae & Felidae & 48.4 & 85.7 & 44.0 & 13.6 & 48.9 & 55.1\\
    \hline
  \end{tabular}
  \caption{Zero-shot performance comparison on AP-10K~\cite{ap-10k}.}
  \label{ap10k-zero-shot}
\end{table*}

\begin{table*}[t]
  \footnotesize
  \centering
  \begin{tabular}{c|c|c|cccc cc}
    \hline
    Method & Backbone & Pre-train & $AP$ & $AP_{50}$ & $AP_{75}$ & $AP_{M}$ & $AP_{L}$ & $AR$ \\
    \hline
    SimpleBaseline~\cite{xiao2018simple} & ResNet-50 & ImageNet  & 68.7 & 93.7 & 76.9 & 63.7 & 69.9 & 73.0 \\
    SimpleBaseline~\cite{xiao2018simple} & ResNet-50 & CLIP      & 70.8 & 94.8 & 79.5 & 67.3 & 72.0 & 75.0\\
    CLAMP (ours)   & ResNet-50 & CLIP      & 72.5 & 94.8 & 81.7 & 67.9 & 73.8 & 76.7\\
    \hline
    SimpleBaseline~\cite{xiao2018simple}   & ViT-Base & CLIP      & 72.3 & 94.7 & 82.1 & 69.4 & 73.3 & 76.3\\
    CLAMP   (ours)  & ViT-Base & CLIP      & 74.3 & 95.8 & 83.4 & 71.9 & 75.2 & 78.3\\
    \hline
  \end{tabular}
  \caption{Performance comparison on Animal-Pose~\cite{cao2019cross}.}
  \label{overall-animal-pose}
\end{table*}

\textbf{Implementation details} We employ the widely used two-stage top-down pose estimation paradigm similar to SimpleBaseline~\cite{xiao2018simple} in the experiments. The ground truth bounding box annotations are utilized in AP-10K to crop animal instances during the training and evaluation process, following the default setting in \cite{ap-10k}. We select the widely used ResNet~\cite{resnet} and the recently introduced attention-based vision transformer ViT~\cite{vit} as the backbone networks for image feature extraction, \ie, ResNet-50 and ViT-Base. We use the ImageNet~\cite{imagenet} pre-trained and CLIP~\cite{clip} pre-trained weights to initialize the backbone networks for evaluating the effect of different pre-training methods. The text encoders after CLIP pre-training, \ie, CLIP-ResNet-50 and CLIP-ViT-Base, are adopted as our language models and initialized with the corresponding CLIP pre-trained weights. The decoder described in SimpleBaseline~\cite{xiao2018simple} is employed to predict the keypoints, which contains several deconvolution layers to upsample the extracted features to $1/4$ of the input resolution and one convolution layer with kernel size $1 \times 1$ to predict the heatmap.

During training, we follow most of the training settings in AP-10K, \ie, the input image of each instance was cropped and resized to $256\times 256$, followed by random flip, rotation, and scale jitter. Each model is trained for a total of 210 epochs with a step-wise learning rate schedule which decays by 10 at the 170th and 200th epoch, respectively. We use the AdamW~\cite{adamw} optimizer with a weight decay of 1e-4. Furthermore, we conduct supervised learning, few-shot learning, and zero-shot learning to thoroughly evaluate the models' performance. For supervised learning on AP-10K and Animal-Pose, we train the model with a batch size of 128 and set 5e-4 as the initial learning rate. An extra learning rate multiplier of 0.1 is applied to the backbone weights to prevent over-fitting when using the ViT model as the backbone. The few-shot and zero-shot learning are conducted on AP-10K as it has much more animal species than Animal-Pose. For few-shot learning, we adopt a batch size of 64 and set the initial learning rate as 5e-4/5e-5 for ResNet-50/ViT-Base. For zero-shot learning configurations, we use a batch size of 128 and set 5e-4 as the initial learning rate. In all experiments, we set $k$ to 8 in Eq.~\ref{eq:prompt} and freeze the text encoder to reduce the computational cost. We show complexity analysis in the supplementary material.

\subsection{Results and analysis}

\subsubsection{Experiments on AP-10K}

\begin{table}[t]
  \centering
  \footnotesize
  \begin{tabular}{l|c}
    \hline
    Method & $AP$ \\
    \hline
    SimpleBaseline~\cite{xiao2018simple} & 69.9 \\
    Hourglass~\cite{newell2016stacked}   & 72.9 \\
    HRNet-w32~\cite{wang2020deep}        & 73.8 \\
    HRNet-w48~\cite{wang2020deep}        & 74.4 \\
    ViPNAS*~\cite{xu2021vipnas}           & 67.1 \\
    HRFormer-S*~\cite{yuan2021hrformer}   & 71.7 \\
    HRFormer-B*~\cite{yuan2021hrformer}   & 73.5 \\
    \hline
    CLAMP-ResNet-50 (ours) & 72.9 \\
    CLAMP-ViT-Base (ours)  & 74.3 \\
    CLAMP-ViT-Large (ours) & 77.8 \\
    \hline
  \end{tabular}
  \caption{Comparison with previous methods in AP-10K~\cite{ap-10k}. * indicates the results using the official mmpose~\cite{mmpose2020} implementation.}
  \label{compare-ap10k}
\end{table}

\textbf{Supervised learning} The results under the supervised learning setting on AP-10K are shown in Table~\ref{overall-ap10k}. It can be observed that the CLIP pre-training can help deliver better results on the animal pose estimation task than using the ImageNet pre-training, \eg, SimpleBaseline~\cite{xiao2018simple} with the CLIP pre-trained ResNet-50 backbone obtains 0.7 AP higher than the counterpart with the ImageNet pre-training. It validates that the image model trained with the prior language knowledge can benefit in dealing with the inter- and intra-species variance in animal pose estimation and thus bringing performance improvements. Furthermore, with the help of the proposed CLAMP, the model achieves much better performance, \ie, there is a performance gain of 2 AP than directly using the CLIP pre-trained model. Such observation demonstrates that with the proposed pose-specific prompts and decomposed adaptation, the language knowledge is better exploited in the animal pose estimation task and brings better performance. Similar conclusion can be drawn by observing the results using the ViT-Base backbone, \eg, the proposed CLAMP outperforms SimpleBaseline by 1.7 AP. Compared with the representative methods~\cite{xiao2018simple, newell2016stacked, wang2020deep} that report results on AP-10K~\cite{ap-10k} and recent pose estimation methods~\cite{xu2021vipnas,yuan2021hrformer}, \ie, the results in Table~\ref{compare-ap10k}, the proposed CLAMP model using ViT-Large as backbone achieves state-of-the-art performance, showing the potential of the proposed CLAMP method, especially given the good scalability of the model size of ViT.

\noindent\textbf{Few-shot learning} We also conduct few-shot learning experiments to study the generalization ability of different methods. We randomly selected 20 samples from each species in the training set of AP-10K to form a 20-shot animal pose estimation training set. The model is tested on the full validation set to evaluate the models' performance under such a challenging training setting with much limited amount of training data available. The results are shown in Table~\ref{ap10k-20shot}. Similar to the observation in supervised learning, CLIP pre-training brings better performance than the visual-only pre-training using ImageNet. For example, SimpleBaseline with a ResNet-50 backbone pre-trained on CLIP outperforms the ImageNet pre-trained counterpart by 0.7 AP (from 51.1 AP to 51.8 AP), further showing the benefit of exploiting language knowledge in animal pose estimation. With the aid of the adaption process, CLAMP outperforms SimpleBaseline by a large margin, \eg, 54.0 AP vs. 51.8 AP with ResNet-50 as the backbone, and 61.2 AP vs. 57.4 AP with ViT-Base as the backbone, respectively. Such observation validates the necessity of adapting the language and visual features to obtain a better performance and enhance the models' generalization ability in the challenging few-shot setting.

\noindent\textbf{Zero-shot learning} We further evaluate the models' generalization ability on unseen animal species in the zero-shot learning experiment. We set up two experimental settings according to whether the animal in the training set and test set belong to the same animal order or not. Since species belonging to the same order have similar appearances while species belonging to different orders have more diverse appearances, these two settings can fully reflect the generalization ability of different methods for dealing with unseen species in different situations. Specifically, we select Bovidae and Canidae as the training and test sets for the different order setting due to their large appearance variance, and Canidae and Felidae as the training and test sets for the same order setting due to their similar appearances. The results are shown in Table ~\ref{ap10k-zero-shot}. It can be observed that compared to the baseline method, the proposed CLAMP model achieves much better performance in both settings, \eg, there is a 5.6 AP and 8.8 AP increase with CLAMP in these two settings, respectively. Such observation well demonstrates that language knowledge can greatly improve the models' generalization ability since the shared language knowledge of keypoints can alleviate the difficulties caused by large visual inter- and intra-species variances.

\subsubsection{Experiments on Animal-Pose}

In addition to AP-10K, we further evaluate the effectiveness of CLAMP on Animal-Pose~\cite{cao2019cross} dataset, which has a different data distribution compared with AP-10K. To train the CLAMP model using the data and annotations from the Animal-Pose dataset, we replace $[KeyPoint]$ in Eq.~\eqref{eq:prompt} with the keypoint names defined in Animal-Pose and expand the number of different keypoints to 20. The training setting is the same as the supervised setting described in the previous section. As shown in Table~\ref{overall-animal-pose}, the benefit of using CLIP pre-training is consistent in both the AP-10K dataset and Animal-Pose dataset, \eg, there is an improvement of 2.1 AP when using the CLIP pre-trained model. By adapting the language knowledge to visual animal pose via the proposed spatial- and feature-level adaption, our CLAMP method with the ResNet-50 and ViT-Base backbones obtains additional 1.7 AP and 2.0 AP gains, respectively. This observation further validates that the introduction of language prompt and exploiting the prior language knowledge can bring general improvements to existing methods in dealing with the large variance in animal pose estimation.

\begin{table}[t]
  \footnotesize
  \centering
  \begin{tabular}{c|c|c|cc}
    \hline
    $\mathcal{L}_{spatial}$ & $\mathcal{L}_{feature}$ & PromptEncoder & $AP$ & $AR$ \\
    \hline
      \ding{53}  & \ding{53}  & \ding{53}   & 70.9 & 74.1\\
      \Checkmark & \ding{53}  & \ding{53}   & 72.1 & 75.3\\
      \Checkmark & \Checkmark & \ding{53}   & 72.6 & 75.8\\
      \Checkmark & \Checkmark & \Checkmark  & 72.9 & 76.3\\
    \hline
  \end{tabular}
  \caption{Ablation study of CLAMP with a CLIP pre-trained ResNet-50 backbone on AP-10K~\cite{ap-10k}.}
  \label{ablation}
\end{table}

\begin{figure}[t]
  \centering
  \includegraphics[width=\linewidth]{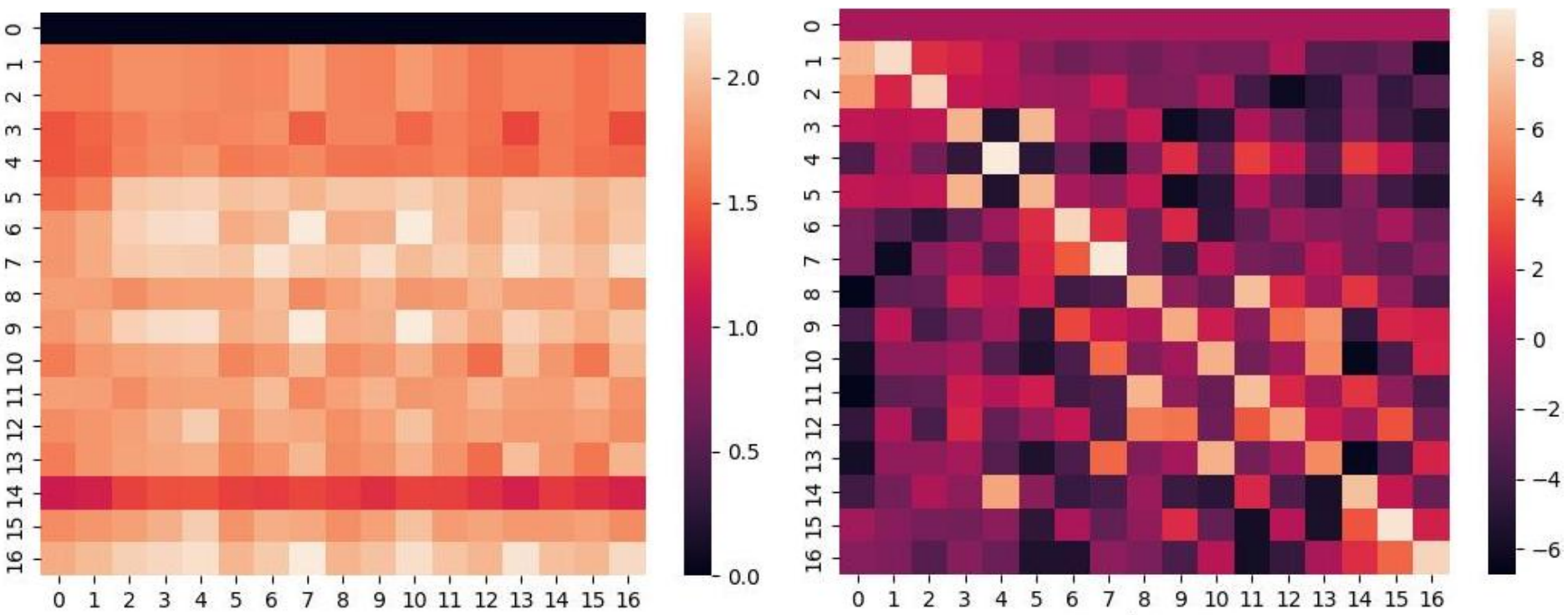}
  \caption{Visualization of the feature-level score map. For each grid-sampled keypoint feature (vertical axis), we calculate its similarity with different text prompts (horizontal axis) after CLIP pre-training (left) and after fine-tuning with the feature-level loss (right). Note that the left eye (\ie, the first keypoint) of the animal is invisible in the test image.
  }
  \label{fig:vis-semantic}
\end{figure}

\subsection{Ablation study}

We use ResNet-50~\cite{resnet} as the backbone network for pose estimation and ablate the effectiveness of the key designs in the CLAMP method in this section, \ie, the spatial-level contrastive loss, the feature-level contrastive loss, and the prompt encoder in pose-specific prompts. The variants are trained under the supervised learning setting for 210 epochs.

\textbf{Spatial-level contrastive loss}
The spatial-level contrastive loss can establish spatial connections between text prompts and image features and provide positional information. To study its impact, we try two variants (\ie, with and without the loss) of the SimpleBaseline with CLIP pre-trained ResNet-50 backbone. As shown in Table~\ref{ablation}, it improves the performance from 70.9 AP to 72.1 AP, validating the benefit of introducing language knowledge into animal pose estimation with the help of spatial-aware adaptation.

\textbf{Feature-level contrastive loss}
As shown in Table~\ref{ablation}, the feature-level contrastive loss further improves the performance by 0.5 AP. The results validate that it is necessary to enhance the discriminative ability of the extracted prompt embeddings and visual features of different keypoints, for a better semantic alignment in animal pose estimation.

\textbf{Prompt encoder}
We also evaluate the effect of the proposed prompt encoder for modeling the relationships between the prompt embeddings. As shown in Table~\ref{ablation}, the prompt encoder further brings a gain of 0.3 AP, \ie, from 72.6 to 72.9, demonstrating the effectiveness of modeling the semantic relationship between descriptions of different keypoints to generate better prompt embeddings and facilitate adapting language knowledge to image features.

\subsection{Visualization and analysis}

\textbf{Feature-level score map} Based on Eq.~\eqref{eq:sementic-contrast}, we can obtain the similarities between text prompts and local keypoint features before and after training with the feature-level loss $\mathcal{L}_{feature}$. As shown in Fig.~\ref{fig:vis-semantic}, the keypoint feature without feature-level adaptation has almost the same similarity for different text prompts, demonstrating the lack of discrimination in the visual keypoint features that are directly extracted by CLIP pre-trained models. With the feature-level adaptation, each keypoint feature has the highest similarity with the corresponding text prompt (\ie, the diagonal elements).
It demonstrates that the feature-level adaptation helps enhance the discrimination of prompt embeddings and visual features of different keypoints, leading to better cross-modal alignment.

\textbf{Spatial-level score map and Qualitative analysis} We visualize the score map to study the effect of the spatial-level contrastive loss in the supplementary material, which shows the established connections between language descriptions and image features and can help understand our CLAMP. In addition, some visual results are presented in the supplementary material, showing the superiority of our method over the baseline model.

\section{Conclusion and discussion}
\label{sec:conclusion} 

This paper proposes CLAMP to introduce prior language knowledge into animal pose estimation. With pose-specific prompts and the spatial-aware and feature-aware adaptation processes, CLAMP provides a promising solution to the well-known challenge in animal pose estimation, \ie, large intra- and inter-species variances together with limited data per species. Extensive experiments on the AP-10K and Animal-Pose benchmarks demonstrate that CLAMP outperforms representative methods by a large margin in all the supervised, few-shot, or zero-shot learning settings. As the first study of cross-modal animal pose estimation, we hope it can provide valuable insights and draw attention from the research community to improve animal pose estimation by effectively exploiting multi-modal knowledge. Besides, the proposed method can also benefit human pose estimation, especially in low-data regimes, which is presented in the supplementary material.

\textbf{Limitation discussion} In this study, we only adopt language models pre-trained on the CLIP dataset, which contains language-image pairs for various scenarios. In the future, we plan to investigate the influence of using an animal pose-related text-image pair dataset for multi-modal pre-training as well as develop more effective visualization tools to explain the learning process and the predictions.

\small\textbf{Acknowledgement} Mr Xu Zhang, Dr Zhe Chen, Mr Yufei Xu, and Dr Jing Zhang were supported by Australian Research Council Projects in part by FL170100117 and IH180100002.


\newpage

{\small
\bibliographystyle{ieee_fullname}
\bibliography{egbib}
}
\newpage
\clearpage
\newpage
\section{Appendix}

In the supplementary material, we show the complexity analysis of our CLAMP and add another ablation study to study the impact of the regularization from losses. Then we evaluate the generalization ability of the proposed CLAMP model by involving more unseen animal species in the zero-shot learning setting. In addition, we visualize the spatial-level score map to further validate the effectiveness of the spatial-level adaptation process in CLAMP by explicitly demonstrating the established spatial connections. At last, we show some pose estimation results from SimpleBaseline~\cite{xiao2018simple} and CLAMP for qualitative comparison, which shows that CLAMP can perform robustly on different animal species with different postures and sizes.

\subsection{Complexity Analysis}
We performed the complexity analysis of our CLAMP and the SimpleBaseline method, and the results are in  Table~\ref{complexity}. The numbers of parameters and GFLOPs are calculated on 256 by 256 images. The results show that our CLAMP brings significant accuracy improvements with a little complexity increase. In practice, the increased complexity is mainly from the cross-attention and prompt encoder. Since our proposed adaptation schemes mainly improve training, they only bring less than 0.01 extra GFLOPs and 0 extra parameters during inference. Note that the prompt embeddings encoded by the text encoder can be stored offline after training and are shared by all the test images, so the text encoder will not add complexity to inference.
\begin{table}[htbp]
  \footnotesize
  \centering
  \begin{tabular}{c|c|c|c|c}
    \hline
    Method & Backbone & Params(M) & GFLOPs & AP\\
    \hline
    SimpleBaseline~\cite{xiao2018simple} & ResNet-50 & 49     & 9.0 & 70.9\\
    CLAMP (ours)     & ResNet-50 & 68     & 9.2  & 72.9\\
    \hline
    SimpleBaseline~\cite{xiao2018simple}   & ViT-Base & 91     & 16.6 & 72.6\\
    CLAMP (ours)      & ViT-Base & 98     & 16.8 & 74.3\\
    \hline
  \end{tabular}
  \caption{Complexity comparison.}
  \label{complexity}
\end{table}

\begin{table}[htbp]
  \footnotesize
  \centering
  \begin{tabular}{c|c|c|c}
    \hline
    Method & Backbone & Embedding type & AP\\
    \hline
    SimpleBaseline~\cite{xiao2018simple}  & ViT-Base & w/o                 & 72.6\\
    CLAMP (ours)   & ViT-Base & trainable matrix     & 72.8\\
    CLAMP (ours)   & ViT-Base & prompt embedding     & 74.3\\
    \hline
  \end{tabular}
  \caption{Ablation study of regularization from losses.}
  \label{ablation-losses}
\end{table}
\begin{table*}[htbp]
  \footnotesize
  \centering
  \begin{tabular}{c|c|c|ccc ccc}
    \hline
    Method & Backbone & Pre-train & $AP$ & $AP_{50}$ & $AP_{75}$ & $AP_{M}$ & $AP_{L}$ & $AR$ \\
    \hline
    SimpleBaseline~\cite{xiao2018simple}   & ViT-Large & CLIP     & 76.9 & 96.0 & 84.4 & 56.5 & 77.2 & 80.0\\
    CLAMP (ours)      & ViT-Large & CLIP     & 77.8 & 96.8 & 85.0 & 58.7 & 78.1 & 81.0\\
    \hline
  \end{tabular}
  \caption{Ablation study for ViT-Large on AP-10K~\cite{ap-10k}.}
  \label{ablation-vitl}
\end{table*}
\begin{table*}[htbp]
  \centering
  \footnotesize
  \begin{tabular}{c|c|c|c|cccccc}
    \hline
    Method & Backbone & Train & Test & $AP$ & $AP_{50}$ & $AP_{75}$ & $AP_{M}$ & $AP_{L}$ & $AR$ \\
    \hline
    SimpleBaseline~\cite{xiao2018simple} & ResNet-50 & Bovidae & Equidae & 41.9 &	71.8 & 40.3	& 27.3 & 42.0 &	46.6\\
    CLAMP (ours)  & ResNet-50 & Bovidae & Equidae & 46.6 &	75.6 & 47.5 & 48.3 & 46.6 &	51.2\\
    \hline
    SimpleBaseline~\cite{xiao2018simple} & ResNet-50 & Bovidae & Felidae & 22.0 &	52.4 & 15.0	& 11.2 & 22.1 &	28.4\\
    CLAMP (ours)  & ResNet-50 & Bovidae & Felidae & 28.7 &	67.6 & 18.9 & 11.9 & 29.0 &	36.3\\
    \hline
    SimpleBaseline~\cite{xiao2018simple}   & ResNet-50 & Canidae & Cricetidae & 16.1 &	41.7 & 10.1 & 3.4 & 16.5 & 26.0\\
    CLAMP   (ours)    & ResNet-50 & Canidae & Cricetidae & 22.0 & 51.1 & 14.1 & 10.1 & 22.7 & 31.6\\
    \hline
    SimpleBaseline~\cite{xiao2018simple}   & ResNet-50 & Canidae & Equidae & 20.5 &	43.9 & 16.6 & 11.1 & 20.7 & 25.3\\
    CLAMP   (ours)    & ResNet-50 & Canidae & Equidae & 28.4 & 59.1 & 23.1 & 9.4 & 29.0 & 34.1\\
    \hline
  \end{tabular}
  \caption{Additional comparisons of the zero-shot generalization performance of different methods on AP-10K~\cite{ap-10k}.}
  \label{ap10k-zero-shot-supp}
\end{table*}
\begin{table*}[htbp]
  \footnotesize
  \centering
  \begin{tabular}{c|c|c|c|c|c|c|c}
    \hline
    Method & Backbone & Pre-train & $AP@1\%$ & $AP@2\%$ & $AP@3\%$ & $AP@5\%$ & $AP@10\%$\\
    \hline
    SimpleBaseline~\cite{xiao2018simple} & ViT-Base & CLIP & 49.5 & 53.3 & 55.5 & 58.4 & 62.1\\
    CLAMP (ours)   & ViT-Base & CLIP & 53.1 & 56.8 & 57.6 & 59.5 & 62.7\\
    \hline
  \end{tabular}
  \caption{Performance on COCO~\cite{coco} in low-data regimes. Note that $AP@k\%$ means the $AP$ on random $k\%$ of COCO data.}
  \label{coco-low-data}
\end{table*}

\subsection{Additional Ablation Study}
We validate that our method can effectively leverage CLIP's capability with the following results. We tested the models' performance under the setting where the $E_{prompt}$ is replaced by a trainable matrix for training, obtaining a performance gain of merely 0.2 AP \textit{w.r.t.} the baseline. Alternatively, our proposed method achieved a gain of 1.7 AP with the help of feature-aware and spatial-aware adaptation. This demonstrates that the major improvements come from our adaptation schemes rather than other components like initialization and regularization from losses. The results are shown in Table~\ref{ablation-losses}.

We also conduct an ablation study for the ViT-Large backbone which has a large number of parameters, and the results can be found in Table~\ref{ablation-vitl}. Even for the significantly large model such as ViT-Large, CLAMP is still effective in taking advantage of the language knowledge for animal pose estimation.

\subsection{Additional Zero-shot Experimental Results}
We report more zero-shot learning experimental results in addition to the results in the main paper to validate the models' generalization ability on unseen animal species, \ie, a) training the model using animals from Bovidae and testing the model using instances from Equidae and Felidae, and b) training the model using animals from Canidae and testing the model using instances from Cricetidae and Equidae. The same training settings as described in the zero-shot experimental setting in the main text are adopted and the results are available in Table~\ref{ap10k-zero-shot-supp}. We can observe that with the help of pose-specific text prompts and the decomposed adaptation process, our CLAMP outperforms the SimpleBaseline by a large margin, \eg, 46.6 AP vs. 41.9 AP, 28.7 AP vs. 22.0 AP, 22.0 AP vs. 16.1 AP, and 28.4 AP vs. 20.5 AP, respectively in these four settings. Such observation validates that language knowledge can improve the model's generalization ability since the shared language knowledge of keypoints can alleviate the difficulties caused by large visual inter- and intra-species variances.

\subsection{Performance on the human pose estimation dataset}
Our method can be extended to human pose by replacing the $KeyPoint$ in Eq.~\eqref{eq:prompt} in the main text with the names of human keypoints, but we found that the human pose estimation methods can easily achieve compelling results by taking advantage of rich labeled data. This can marginalize the benefits of the knowledge in CLIP. In practice, we have tested our method on the COCO human pose dataset~\cite{coco}, obtaining 0.4 AP improvement. However, we found that our CLAMP can bring more benefits for human pose estimation in low-data regimes. We test our method on random $k\%$ of COCO data for $k \in \{1, 2, 3, 5, 10\}$ and display the mean results of three times random sampling on COCO in Table.~\ref{coco-low-data}.

\subsection{Visualization of Spatial-level Score Maps}
\begin{figure}[htbp]
  \centering
  \includegraphics[width=\linewidth]{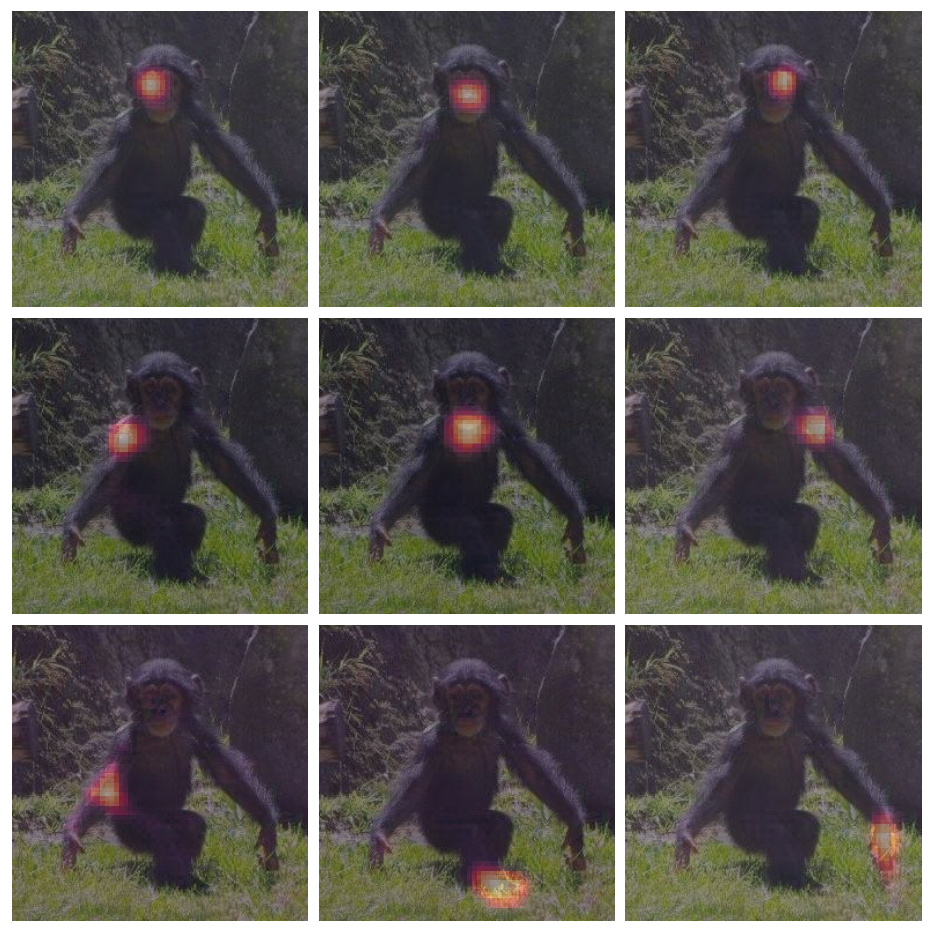}
  \caption{Visualization of the spatial-level score maps. In order to intuitively display the relationship between the keypoint presence score and the animal image, we superimpose the obtained score maps on the animal image. As shown in the figure, the lightness and darkness of the pixels are used to indicate the keypoint presence score (the brighter pixels indicate the higher scores). The superimposed images for different keypoints are displayed in the following order: right eye, nose, left eye, right shoulder, neck, left shoulder, right elbow, right front paw, and left front paw (from the first row to the last row, from left to right in each row).
  }
  \label{fig:pixel-vis}
\end{figure}

Based on Eq.~\eqref{eq:pixel-1} and Eq.~\eqref{eq:pixel-2} in the main text, we can obtain the keypoint presence score on each spatial position of the input image with the help of spatial-level loss $\mathcal{L}_{spatial}$. We visualize the upsampled score maps in Fig.~\ref{fig:pixel-vis}, which displays the established spatial connections between language descriptions and image features in our CLAMP. It can be seen that for each keypoint description, the highest score values show up in the corresponding image region that has the same semantics as the keypoint description. This indicates that the spatial-level loss helps establish spatial connections between language knowledge and visual features, which can provide positional information for animal poses.

\subsection{Qualitative Analysis}

\begin{figure*}[htbp]
  \centering
  \includegraphics[width=\linewidth]{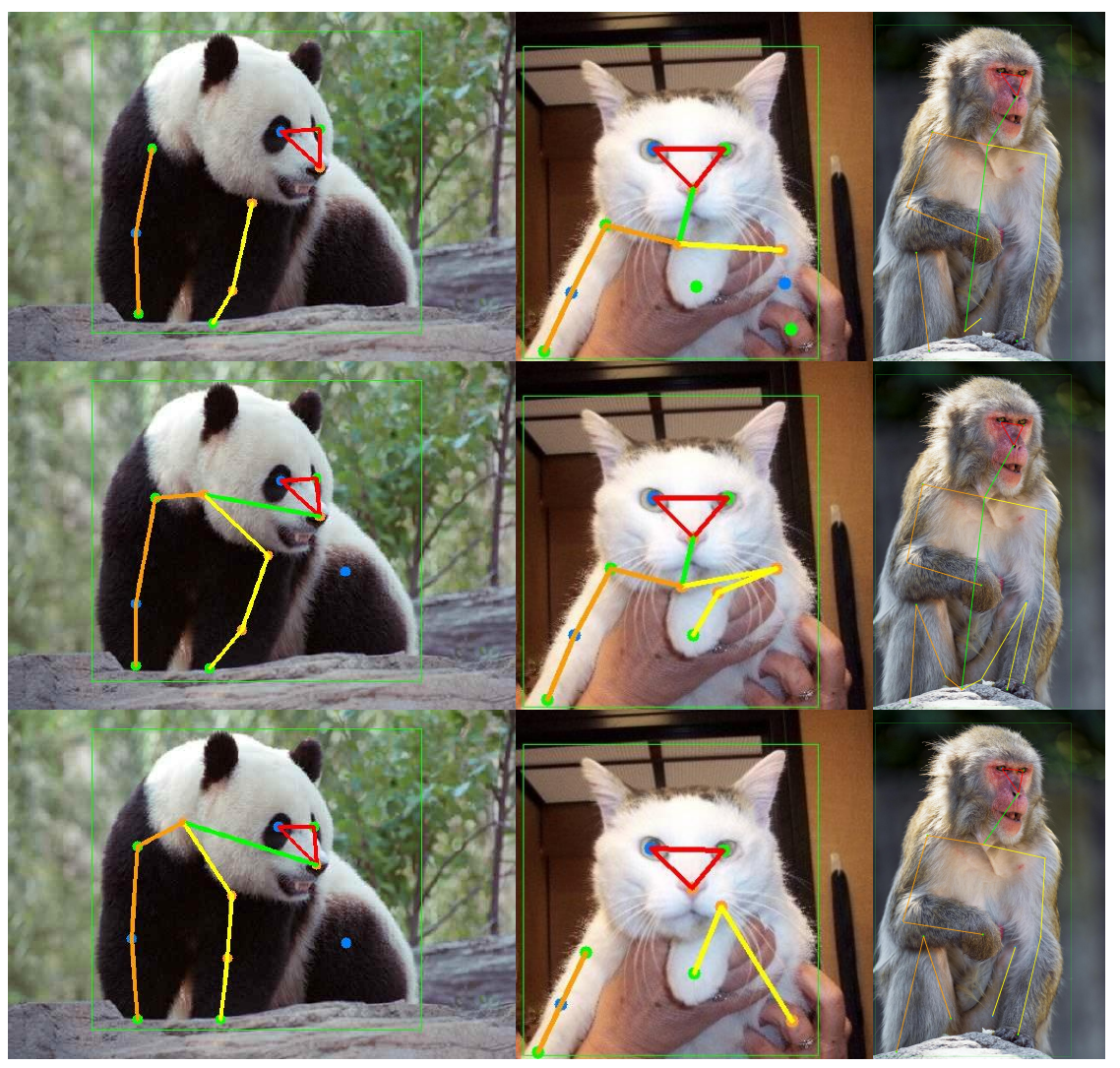}
  \caption{Qualitative analysis of the SimpleBaseline ~\cite{xiao2018simple} (the first row) and the proposed CLAMP (the second row). The ground truth poses are shown in the last row.
  }
  \label{fig:qua1}
\end{figure*}

\begin{figure*}[htbp]
  \centering
  \includegraphics[width=\linewidth]{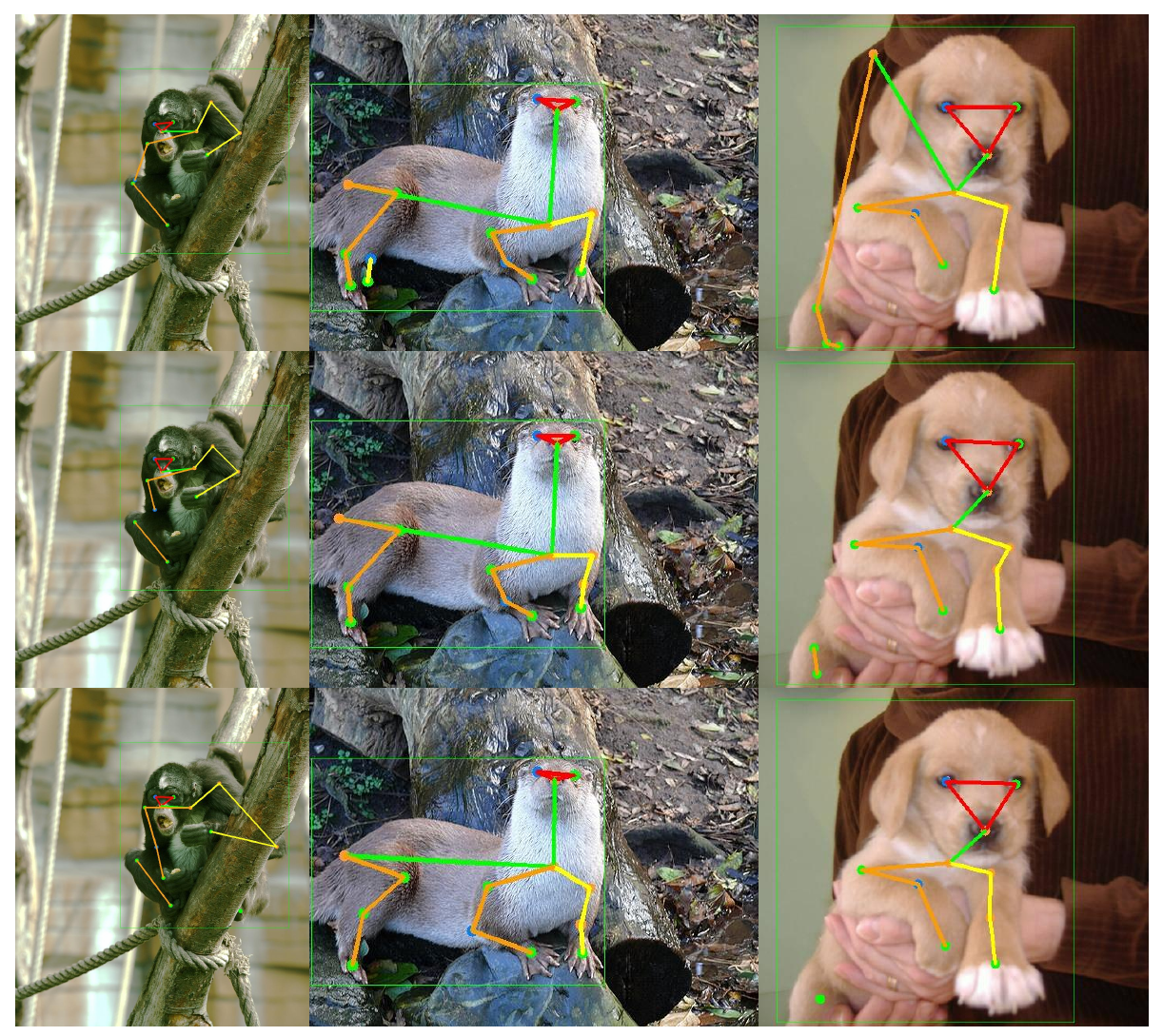}
  \caption{More qualitative results of the SimpleBaseline ~\cite{xiao2018simple} (the first row) and the proposed CLAMP (the second row). The ground truth poses are shown in the last row.
  }
  \label{fig:qua2}
\end{figure*}

To get an intuitive understanding of the proposed method, we show some qualitative results in Fig.~\ref{fig:qua1} and Fig.~\ref{fig:qua2}. In each figure, the baseline method, \ie, SimpleBaseline~\cite{xiao2018simple}, the proposed CLAMP, and the ground truth are shown from top to bottom. 
As can be seen, our method can produce accurate pose estimation results on animals with large variances in appearances and poses. 
Taking the first column in Fig.~\ref{fig:qua1} as an example, the baseline method in the first row overlooks the neck and left hip of the panda. By contrast, our CLAMP successfully leverages the language knowledge and outputs all keypoints that are labeled in the ground truth. 
Similar results can also be observed in the images of other animal species. 
Besides, CLAMP can sometimes produce more accurate results than human annotations. For example, in the second column, the neck, the left elbow, and the left shoulder are neglected or probably incorrectly labeled. By contrast, our CLAMP can locate and recognize those keypoints, demonstrating its potential in dealing with hard cases.
\end{document}